\documentclass[11pt,a4paper]{article}
\usepackage[hyperref]{emnlp2020}
\usepackage{times}
\usepackage{latexsym}

\usepackage{microtype}

\usepackage{soul}
\usepackage{url}
\usepackage[utf8]{inputenc}
\usepackage{caption}
\usepackage{graphicx}
\usepackage{amsthm}
\usepackage{booktabs}
\usepackage{algorithm}
\usepackage{algorithmic}
\usepackage{multirow}
\usepackage{diagbox}
\usepackage{color}
\usepackage{xcolor}
\usepackage{bm}
\usepackage{CJKutf8}
\usepackage{subfig}
\usepackage{amsmath,amssymb,amsfonts}

\aclfinalcopy 

\title{HENIN: Learning Heterogeneous Neural Interaction Networks\\for Explainable Cyberbullying Detection on Social Media}

\author{Hsin-Yu Chen \\
  Institute of Data Science \\
  National Cheng Kung University \\
  Tainan, Taiwan \\
  \texttt{d0107330@gmail.com} \\\And
  Cheng-Te Li \\
  Institute of Data Science \\
  National Cheng Kung University \\
  Tainan, Taiwan \\
  \texttt{chengte@ncku.edu.tw} \\}

\date{}

\begin{document}
\maketitle
\begin{abstract}
In the computational detection of cyberbullying, existing work largely focused on building generic classifiers that rely exclusively on text analysis of social media sessions. Despite their empirical success, we argue that a critical missing piece is the model explainability, i.e., why a particular piece of media session is detected as cyberbullying. In this paper, therefore, we propose a novel deep model, HEterogeneous Neural Interaction Networks (HENIN), for explainable cyberbullying detection. HENIN contains the following components: a comment encoder, a post-comment co-attention sub-network, and session-session and post-post interaction extractors. Extensive experiments conducted on real datasets exhibit not only the promising performance of HENIN, but also highlight evidential comments so that one can understand why a media session is identified as cyberbullying.
\end{abstract}

\section{Introduction}
\begin{figure}[!t]
  \centering
  \includegraphics[width=0.9\linewidth]{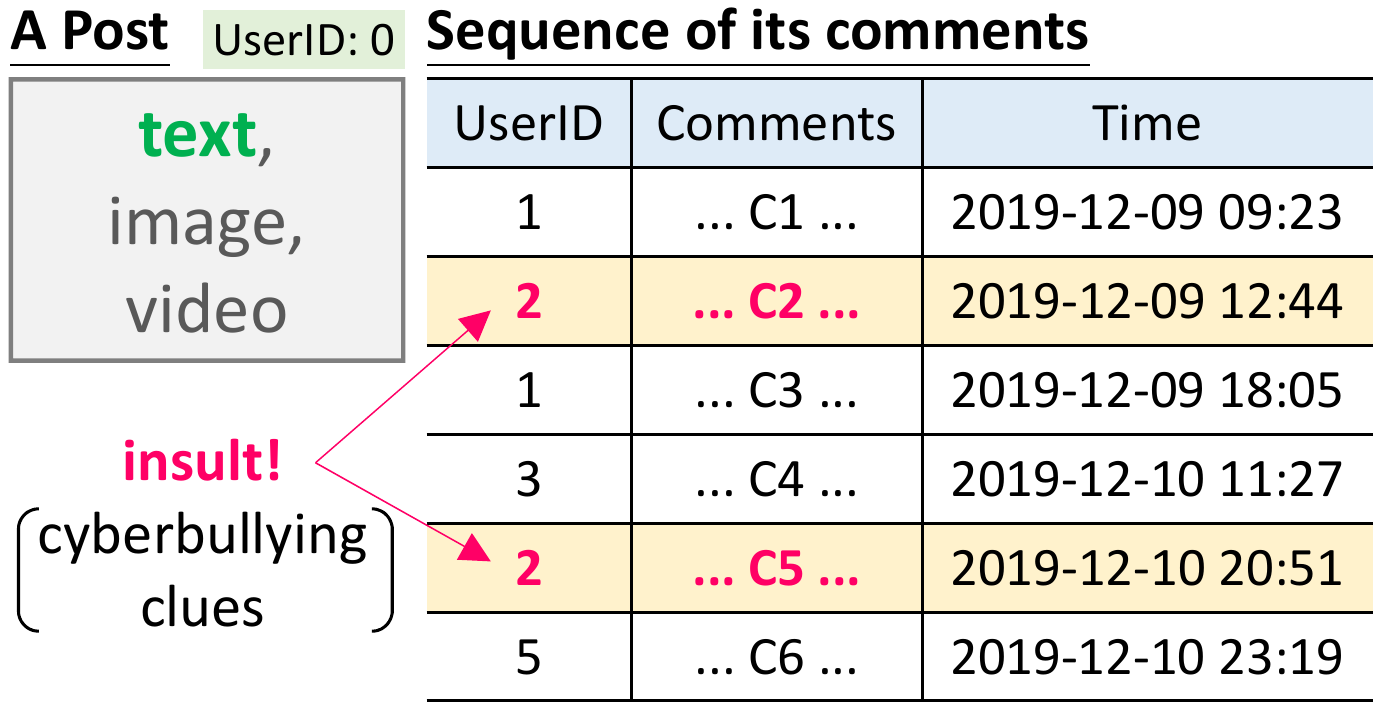}
  \caption{An illustration of a media session containing an image/video/posted text and a sequence of comments. A cyberbullying session is typically composed of multiple insulting comments.}
  \label{fig:session-illustration}
  \vspace{-8pt}
\end{figure}

In recent years, cyberbullying has become one of the most pressing online risks among youth and raised serious concerns in society. Cyberbullying is commonly defined as the electronic transmission of insulting or embarrassing comments, photos or videos, as illustrated in Figure~ \ref{fig:session-illustration}. Harmful bullying behavior can include posting rumors, threats, pejorative labels, and sexual remarks. Research from the American Psychological Association and the White House has revealed more than $40\%$ of young people in the US indicate that they have been bullied on social media platforms~\citep{dinakar2012common}. Such a growing prevalence of cyberbullying on social media has detrimental societal effects, such as victims may experience lower self-esteem, increased suicidal ideation, and a variety of negative emotional responses~\citep{hinduja2014bullying}. Therefore, it has become critically important to be able to detect and prevent cyberbullying on social media. Research in computer science aimed at identifying, predicting, and ultimately preventing cyberbullying through better understanding the nature and key characteristics of online cyberbullying. 

In the literature, existing efforts toward automatically detecting cyberbullying have primarily focused on textual analysis of user comments, including keywords~\citep{dadvar2012improved,nahar2013effective,nand2016bullying} and sentiments analysis ~\citep{dani2017sentiment}. These studies attempt to build a generic binary classifier by taking high-dimensional text features as the input and make predictions accordingly. Despite their satisfactory detection performance in practice, these models largely overlooked temporal information of cyberbullying behaviors. They also ignore user interactions in social networks. Furthermore, the majority of these methods focus on detecting cyberbullying sessions effectively but cannot explain ``why'' a media session was detected as cyberbullying. Given a sequence of comments with user attributes, we think sequential learning can allow us to better exploit and model the evolution and correlations among individual comments. Besides, graph-based learning can enable us to represent and learn how users interact with each other in a session. 

This work aims to detect cyberbullying by jointly exploring explainable information from user comments on social media. To this end, we build an explainable cyberbullying detection framework, \underline{\textbf{HE}}terogeneous \underline{\textbf{N}}eural \underline{\textbf{I}}nteraction \underline{\textbf{N}}etworks (\textbf{HENIN}), through a coherent process. HENIN consists of three main components that learn various interactions among heterogeneous information displayed in social media sessions. A comment encoder is created to learn the representations of user comments through a hierarchical self-attention neural network so that the semantic and syntactic cues on cyberbullying can be captured. We create a post-comment co-attention mechanism to learn the interactions between a posted text and its comments. Moreover, two graph convolutional networks are leveraged to learn the latent representations depicting how sessions interact with one another in terms of users, and how posts are correlated with each other in terms of words.

Specifically, we address several challenges in this work: (a) how to perform explainable cyberbullying detection that can boost detection performance, (b) how to highlight explainable comments without the ground truth, (c) how to model the correlation between posted text and user comments, and (d) how to model the interactions between sessions in terms of users, and the interactions between textual posts in terms of words. Our solutions to these challenges result in a novel framework HENIN. 

Our contributions are summarized as follows.
(1) We study a novel problem of explainable cyberbullying detection on social media.
(2) We provide a novel model, HENIN~\footnote{The Code of HENIN model is available at: \url{https://github.com/HsinYu7330/HENIN}}, which jointly exploits posted text, user comments, and the interactions between sessions and between posts to learn the latent representations for cyberbullying detection.
(3) Experiments conducted on Instagram and Vine datasets exhibit the promising performance of HENIN, and the evidential comments and words highlighted by HENIN, for detecting cyberbullying media sessions with explanations.

\section{Related Work}
Relevant studies can be categories into social contexts-based and user comment-based approaches.
\textbf{Social contexts-based approaches} utilize three categories of features, user-based, post-based, and network-based. 
(a) Post-based features rely on text analysis to identify cyberbullying evidences (e.g., profane words) on social media~\citep{dadvar2012improved,nahar2013effective,nand2016bullying}. 
\citet{xu2012learning} point out Latent Semantic Analysis(LSA) and Latent Dirichlet Allocation (LDA) can be used to learn latent representations of posts. In addition, \textit{SICD}~\citep{dani2017sentiment} further models post sentiments for cyberbullying detection. 
(b) User-based features are extracted from user profiles to measure their characteristics. Gender-specific features, user's past posts, account registration time, and frequently-used words are useful user-based features~\citep{dadvar2012cyberbullying,dadvar2013improving}.
(c) Existing studies~\citep{cheng2019xbully,tu2018structural,wang2017signed} also prove that network-based features are effective in detecting cyberbullying. These features are learned by constructing propagation networks or interaction networks that depict how posts are spread and how users interact with each other. 
\textbf{User comment-based approaches} utilize the sequence of user comments to detect cyberbullying of the source post.
CONcISE~\citep{yao2019cyberbullying} is a sequential hypothesis testing method conducted on the comment sequence to select the significant comment features.
\citet{raisi2018weakly} detect harassment-based cyberbullying by identifying expert-provided key phrases from user comments.

\section{Problem Statement}
Let $S=\{s_1, s_2,..., s_M\}$ denote a corpus of $M$ social media sessions. Each media session contains the posted text and its subsequent comments. Let $P$ be a posted text, consisting of $N$ words $\{w_i\}^N_{i=1}$. Let $C=\{c_1, c_2,..., c_T\}$ be a set of $T$ comments related to the post $P$, where each comment $c_j=\{w^j_1, w^j_2,...w^j_{Q_j}\}$ contains $Q_j$ words. Let $G_{ss}=(V_S,E_S)$ be a session-session weighted graph, in which we consider each media session as a node $s\in V_S$ and the similarity between sessions as an edge weight $e_{(s_i, s_j)}\in E_S$. Let $G_{pp}=(V_P,E_P)$ be a post-post weighted graph, in which we consider each posted text as a node $p\in V_P$ and the similarity between posts as an edge weight $e_{(p_i, p_j)}\in E_P$. We treat the cyberbullying detection problem as the binary classification problem, i.e., each media session is associated with a binary label $y=\{0, 1\}$ with 1 representing a bullying session, and 0 representing a non-bullying session. At the same time, we aim to learn a rank list $RC$ from all comments in $\{c_j\}^T_{j=1}$, according to the degree of explainability, where $RC_k$ denotes the $k_{th}$ most explainable comment. The explainability of comments denotes the impact degree of detecting the media session is cyberbullying or not. Formally, we can represent the problem as \textit{Explainable Cyberbullying Detection}.

\textbf{Problem:} Given a posted text $P$, a set of related comments $C$, the session graph $G_{ss}$ and the post graph $G_{pp}$, the goal is to learn a cyberbullying detection function $f: f(P, C, G_{ss}, G_{pp})\rightarrow (\hat{y}, RC)$, such that it maximizes the prediction accuracy with explainable comments ranked highest in $RC$.

\section{The proposed HENIN Model}
\begin{figure}[!t]
  \centering
  \includegraphics[width=1.0\linewidth]{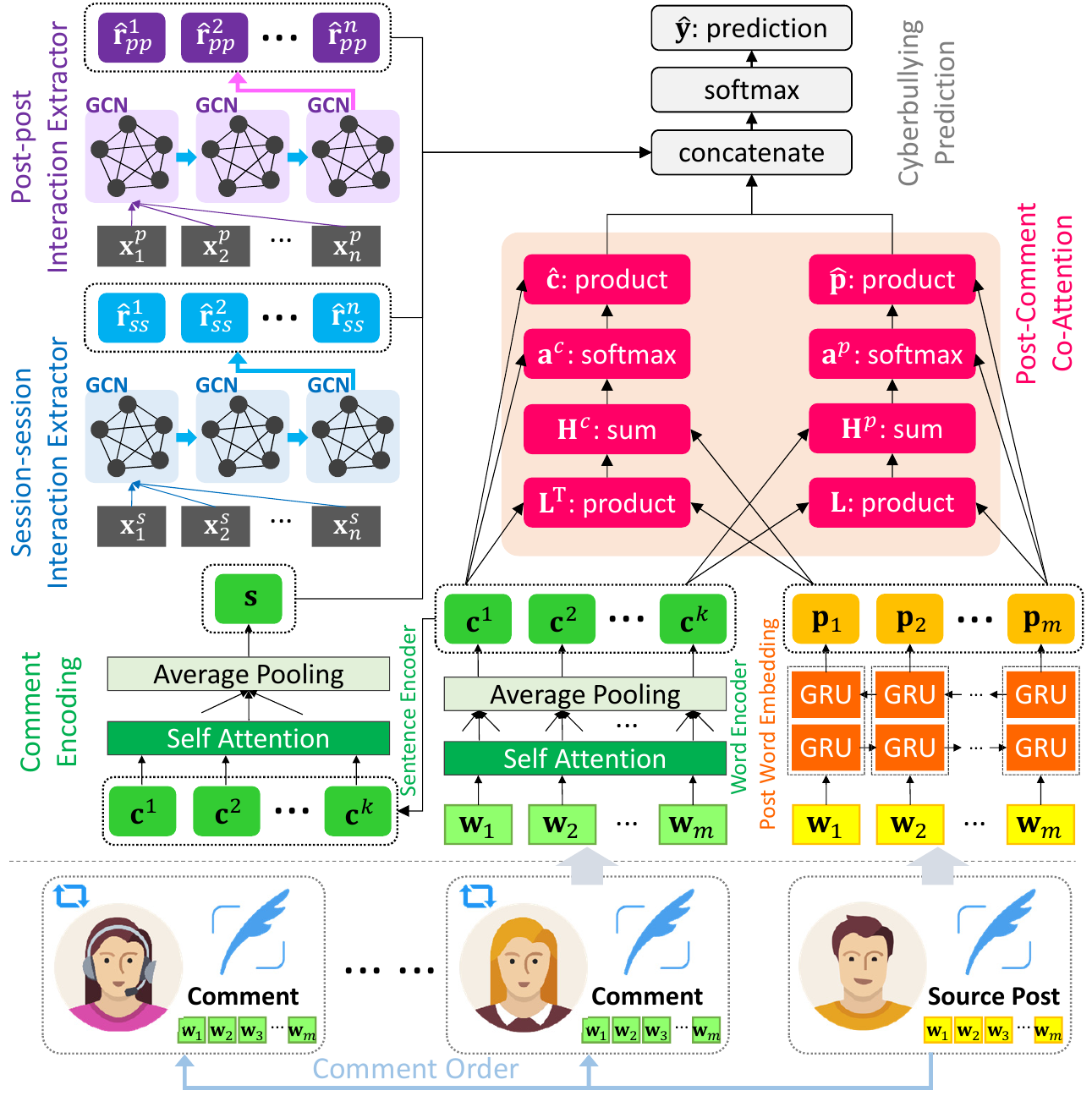}
  \caption{The proposed HENIN model, which contains four components: a joint word-level and sentence-level comment encoder, a post-comment co-attention mechanism, session-session and post-post interaction extractors, and the final cyberbullying prediction.}
  \label{fig:framework}
\end{figure}

In this section, we present the details of the proposed HENIN, which jointly learns the hierarchical self-attention and graph convolutional neural networks for cyberbullying detection. It consists of four major components (Figure~\ref{fig:framework}): (1) a comment encoder (including word-level and sentence-level), (2) a post-comment co-attention mechanism, (3) session-session and post-post interaction extractors, and (4) a cyberbullying prediction component.

The comment encoder component depicts the modeling from the comment linguistic features to latent representation features through hierarchical word-level and sentence-level self-attention networks. The explainability degree of comments is learned through the attention weights within sentence-level self-attention learning. The post-comment co-attention mechanism is performed in the level of word embeddings. The mutual interactions between the posted text and comments can be learned through the post-comment co-attention. On the other hand, the session-session interaction extractor and the post-post interaction extractor aim at modeling how users interact across media sessions, and how words are correlated across posts, through two graph convolutional neural networks. Finally, the cyberbullying prediction is made by concatenating the representations of the aforementioned three components.

\subsection{Comment Encoding}
A set of comments related to the given media session contains linguistic cues at the word and sentence levels. Textual usages in comments provide different degrees of importance for explainability of why the session is detected as cyberbullying. For example, in a cyberbullying media session extracted from the Instagram dataset (see Section~\ref{sec:data}), the comment \textsl{``how the fuck are you even a fucking fan you cunt if you just talk shit about harry fuck you kaitlyn!''}, the words ``fuck'' and ``shit'' contribute more signals to reflect apparent and evidential emotion sense, compared to other ones. Meanwhile, this comment strongly expresses malicious remarks to someone, and therefore it is not only more explainable but also useful to determine whether it is a cyberbullying session. 

Several studies have shown that improved document representations with highlighting important words and sentences for classification can be learned by hierarchical attention neural networks~\citep{yang2016hierarchical,cheng2019hierarchical}. Inspired by~\citep{yang2016hierarchical}, we adopt a hierarchical neural network to model word-level and sentence-level representations through self-attention mechanisms. Specifically, we first learn the comment embedding vector by utilizing the word encoder with self-attention. Then we learn the comment representations through the sentence encoder with self-attention.

\paragraph{Word Encoder.}  
Given a comment $c_j$ with $m$ words, we first embed the words to a latent space via the pre-trained word2vec model~\citep{mikolov2013distributed}. Then we capture words' contextual relations among comments by calculating scaled dot-product attention~\citep{vaswani2017attention}. Specifically first, let word embeddings as input vectors $\mathbf{x}_i$. The query vector sequence $\mathbf{q}_i$, the key vector sequence $\mathbf{k}_i$, and the value vector sequence $\mathbf{v}_i$ can be obtained by linear transformation, i.e., $\mathbf{q}_i=\mathbf{w}_q\mathbf{x}_i$, $\mathbf{k}_i=\mathbf{w}_k\mathbf{x}_i$, and $\mathbf{v}_i=\mathbf{w}_v\mathbf{x}_i$, where $\mathbf{w}_q, \mathbf{w}_k, \mathbf{w}_v$ are the learnable parameters through the networks. Next we compute the dot products of the query with all keys, divide each by $\sqrt{d_k}$ ($d_k$ is the dimension of keys), and apply a softmax function to obtain the attention weights on the values:
$
\mathbf{a}_i=\text{softmax}(\frac{\mathbf{q}_i\mathbf{k}_i^\top}{\sqrt{d_k}})
$,
where $\mathbf{a}_i$ is an attention weight vector that measures the importance of each word in the comment. Finally, each word's hidden representation can be obtained by computing the dot products of attention weights $\mathbf{a}_i$ and the value vector sequence $\mathbf{v}_i$. We take the average of the learned representations to generate the comment vector $\mathbf{c}^j$, given by:
$
\mathbf{c}^j = \frac{\sum_{i=1}^{m}\mathbf{a}_i\mathbf{v}_i}{m}
$.

\paragraph{Sentence Encoder.} Similar to the word encoder, we utilize the scaled dot-product attention to encode each media session. The aim is to capture the context information at the sentence level, and to generate the media session representation of post $P_i$, denoted by $\mathbf{s}^i$, from the learned comment embedding vectors $\{\mathbf{c}^1, \mathbf{c}^2,..., \mathbf{c}^k\}$. Every post's sentence embedding $\mathbf{s}$ will be used as features for cyberbullying prediction.

\subsection{Post-Comment Co-attention Mechanism}
To model the interaction between posted text and comments, we propose a post-comment co-attention mechanism that learns the semantic word-level correlation between posted text and comments. That said, we intend to simultaneously learn and derive the attention weights of words on posted text and comments. Specifically first, similar to comment encoding, word embeddings of a posted text are obtained by a pre-trained word2vec model. We adopt recurrent neural networks with bidirectional gated recurrent units (GRU) to model word sequences from both directions of words. The bidirectional GRU contains the forward GRU $\overrightarrow{f}$ that reads posted text $p^i$ from word $w^i_1$ to $w^i_m$ and the backward GRU $\overleftarrow{f}$ that reads posted text $p^i$ from word $w^i_m$ to $w^i_1$, given by:
$
\overrightarrow{\mathbf{h}^i_t}=\overrightarrow{GRU}(\mathbf{w}^i_t) (t\in \{1,..., m\})$ and $
\overleftarrow{\mathbf{h}^i_t}=\overleftarrow{GRU}(\mathbf{w}^i_t) (t\in \{m,..., 1\})
$.
We obtain the embedding of word $p^i_t$ in a posted text by concatenating its forward and backward hidden states $\overrightarrow{\mathbf{h}^i_t}$ and $\overleftarrow{\mathbf{h}^i_t}$, i.e., $\mathbf{p}^i_t=[\overrightarrow{\mathbf{h}^i_t}, \overleftarrow{\mathbf{h}^i_t}]$. Then we can construct the feature matrix of words of posted text $\mathbf{P}=[\mathbf{p}^1,..., \mathbf{p}^N]$. Similarly the feature matrix of comments $\mathbf{C}=[\mathbf{c}^1,..., \mathbf{c}^T]$ can be derived. 

The proposed co-attention mechanism attends to the posted text words and the comment simultaneously. By extending the co-attention formulation~\citep{lu2016hierarchical,cui2019defend}, we first compute the affinity matrix $\mathbf{L}\in \mathbb{R}^{T \times N}$: 
$
\mathbf{L} = \text{tanh}(\mathbf{C}^{\top}\mathbf{W}_l\mathbf{P})
$,
where $\mathbf{W}_l$ is a matrix of learnable weights. The affinity matrix $\mathbf{L}$ is used to transform the comment attention space to the posted text attention space, and vice versa for $\mathbf{L}^\top$. As a result, we can consider the affinity matrix as a feature matrix, and learn to predict the posted text and comment attention maps $\mathbf{H}^p$ and $\mathbf{H}^c$, as follows:
$
\mathbf{H}^p=\text{tanh}(\mathbf{W}_p\mathbf{P}+(\mathbf{W}_c\mathbf{C})\mathbf{L})$
, and $
\mathbf{H}^c=\text{tanh}(\mathbf{W}_c\mathbf{C}+(\mathbf{W}_p\mathbf{P})\mathbf{L}^\top)
$,
where $\mathbf{W}_p, \mathbf{W}_c$ are the matrices of learnable parameters. The attention weights of posted text and comments, $\mathbf{a}^p$ and $\mathbf{a}^c$, can be obtained by:
$
\mathbf{a}^p = \text{softmax}(\mathbf{w}^\top_{hp}\mathbf{H}^p)
$,
$
\mathbf{a}^c = \text{softmax}(\mathbf{w}^\top_{hc}\mathbf{H}^c)
$,
where $\mathbf{w}^\top_{hp}$ and $\mathbf{w}^\top_{hc}$ are vectors of learnable weight parameters. Based on the above attention weights, the posted text and comment attention vectors are obtained by calculating the weighted sum of the posted text features and comment features via: 
$
\hat{\mathbf{p}}=\sum_{i=1}^{N}\mathbf{a}^p_i\mathbf{p}^i
$ and $\hat{\mathbf{c}}=\sum_{i=1}^{T}\mathbf{a}^c_i\mathbf{c}^i, 
$,
where $\hat{\mathbf{p}}$ and $\hat{\mathbf{c}}$ are the learned features vectors for posted text and comments, respectively, through the co-attention mechanism.

\subsection{Interaction Extractors}
To learn and represent the potential interactions between two sessions as well as two text posts, we utilize multilayer neural networks that operate on graph data based on the layers of graph convolutional networks (GCN)~\citep{kipf2016semi}. GCN is able to induce embedding vectors of nodes based on features of their neighborhoods. We create two multi-layer GCNs to learn the embeddings of the given session $s_i$ and its posted text $P_i$ from the session-session graph $G_{ss}$ and the post-post graph $G_{pp}$, respectively.

\textbf{Session-session Interaction Extractor.} Let $\mathbf{X}=(\mathbf{x}_1, \mathbf{x}_2,...,\mathbf{x}_n) \in \mathbb{R}^{n\times p}$ be the vectors of user participation in all sessions, where $n$ is the number of all sessions and $p$ is the number of users. Each vector $\mathbf{x}_i$ is a multi-hot encoding that depicts how session $s_i$ is participated by all users. Let matrix $\hat{\mathbf{R}}_{ss}$ be the representations of all sessions learned from the session-session graph $G_{ss}=(\mathbf{X}, \mathbf{A})$, where $\mathbf{A}\in \mathbb{R}^{n \times n}$ encodes the pairwise relationships (such as cosine similarity, which is used by default) between sessions. We exploit GCN to learn $\hat{\mathbf{R}}_{ss}$. GCN contains one input layer, several propagation layers, and the final output layer~\citep{kipf2016semi}. At deeper layers, the nodes indirectly receive more information from farther nodes in the graph. Given the input feature matrix $\mathbf{X}^{(0)}=\mathbf{X}$ and the graph structure matrix $\mathbf{A}$, GCN performs the layer-wise propagation in hidden layers via 
$
\mathbf{X}^{(k+1)}=\rho(\hat{\mathbf{A}}\mathbf{X}^{(k)}\mathbf{W}^{(k)})
$,
where $k=0, 1,..., K-1$ and $\mathbf{W}^{(k)}$ is the matrix of learnable parameters in the $k$-th layer. $\rho$ is a non-linear activation function, such as ReLU, and $\mathbf{X}^{(k+1)}$ denotes the activation output in the $k$-th layer. $\hat{\mathbf{A}}$ is the normalized symmetric adjacency matrix, $\hat{\mathbf{A}}=\mathbf{D}^{-\frac{1}{2}}\mathbf{A}\mathbf{D}^{-\frac{1}{2}}$, where $\mathbf{D}=diag(d_1, d_2,..., d_n)$ is a diagonal matrix with $d_i=\sum_{j=1}^{n}\mathbf{A}_{ij}$. Finally, the graph representations $\hat{\mathbf{R}}_{ss}=[\hat{\mathbf{r}}_{ss}]$ can be obtained from the output layer that uses \textit{softmax} as the activation function.

\textbf{Post-post Interaction Extractor.} Similar to \textit{session-session interaction extractor}, we depict each posted text in the graph $G_{pp}$ as a real-valued vector $\mathbf{x}_i$ by using the word embedding vector of post $P_i$ as the initial feature. By performing GCNs as aforementioned, we can derive the graph representations of all posts, denoted by $\hat{\mathbf{R}}_{pp}=[\hat{\mathbf{r}}_{pp}]$.

\subsection{Cyberbullying Prediction}
By concatenating the sentence embedding vector $\mathbf{s}$, the post-comment co-attention feature vectors $\hat{\mathbf{p}}$ and $\hat{\mathbf{c}}$, the session interaction representation $\hat{\mathbf{r}}_{ss}$, and the post interaction representation $\hat{\mathbf{r}}_{pp}$, we generate the prediction via a fully-connected layer, given by:
$
\hat{\mathbf{y}}=\sigma([\hat{\mathbf{p}}, \hat{\mathbf{c}}, \mathbf{s}, \hat{\mathbf{r}}_{ss}, \hat{\mathbf{r}}_{pp}]\mathbf{W}_f+\mathbf{b}_f)
$,
where $\hat{\mathbf{y}}$ is the predicted probability vector indicating the predicted probability of label 1 (i.e., cyberbullying). $\mathbf{W}_f$ and $\mathbf{b}_f$ are the learnable parameters and biases. $\sigma$ is the sigmoid function.  $y\in \{0, 1\}$ denotes the ground-truth label of media sessions. The goal is to minimize the cross-entropy loss function:
$
\mathcal{L}(\Theta)=-y\log(\hat{\mathbf{y}})-(1-y)\log(1-\hat{\mathbf{y}})
$,
where $\Theta$ denotes all parameters of the network. The parameters in the network are learned through the \textit{Adam} optimizer~\citep{kingma2014adam}, which is an adaptive learning rate method that uses estimations of first and second moments of gradient to adapt the learning rate for each weight of the neural network. We choose \textit{Adam} since it is generally regarded as being fairly robust and effective to the choice of the hyperparameters, and it is widely used for training neural networks.

\section{Experiments}
We aim to answer the following evaluation questions.
\textbf{EQ1:} Can HENIN improve the cyberbullying media session \textit{classification performance}?
\textbf{EQ2:} How effective is \textit{each component} of HENIN? 
\textbf{EQ3:} Is HENIN able to perform accurate \textit{early} detection of cyberbullying sessions?
\textbf{EQ4:} Can HENIN highlight comments that can \textit{explain why} a media session is detected as cyberbullying?

\subsection{Datasets and Settings}
\label{sec:data}
We use two social media datasets whose statistics is shown in Table~\ref{tab:data-stat}. One is Instagram dataset~\citep{hosseinmardi2015analyzing,hosseinmardi2016prediction}, which contains image description and user comments. The other is Vine~\citep{rafiq2015careful,rafiq2016analysis}, which is a mobile application website that allows users to record and edit a few seconds looping videos. The texts of both datasets are in English.

\begin{table}[!t]
    \caption{Statistics of Instagram and Vine datasets.}
    \centering
    \begin{tabular}{l|r|r}
    \hline
    Datasets & Instagram & Vine \\ \hline
    \# Sessions & 2,211 & 882 \\
    \# Bullying & 676 & 283 \\
    \# Non-Bullying & 1,535 & 599 \\
    \# Comments & 159,277 & 70,385 \\
    \# Users & 72,176 & 25,699 \\ \hline
    \end{tabular}
    \label{tab:data-stat}
\end{table}

We compare our HENIN model with several methods, including classification models such as Logistic Regression (\textbf{LR})~\citep{hosseinmardi2015analyzing,hosseinmardi2016prediction} and Random Forest (\textbf{RF})~\citep{rafiq2015careful,rafiq2016analysis}. We collect posted text and all related comments of the session as a document to embed the session to a latent space via pre-trained doc2vec model~\citep{le2014distributed}. Then we leverage the session representations as input features to train LR and RF classifiers. In addition, we also compare HENIN with three end-to-end deep learning models, including \textbf{RNN}, \textbf{GRU}, and GRU with attention \textbf{GRU+A}. We also compare HENIN with a recent advance \textbf{CONcISE}~\citep{yao2019cyberbullying}, which has a sequential hypothesis testing-based mechanism to produce timely and accurate detection of cyberbullying. For a fair comparison with CONcISE, we follow their settings by using their suggested key terms: ``ugly'', ``shut'', ``suck'', ``gay'', ``beautiful'', ``sick'', `bitch'', `work'', ``hate'', and ``fuck.''

We provide the hyperparameter settings to enable the reproducibility. (1) The maximum number of words per comment \texttt{MAX\_COM\_WORD\_LEN=10} and \texttt{6} on Instagram and Vine, respectively, according to the median of all comments' length. (2) The maximum length of user comments \texttt{MAX\_COM\_LEN=75} and \texttt{80} on Instagram and Vine, respectively. (3) The dimension of word embeddings \texttt{d=300}. (4) The number of GCN layers is \texttt{3}. (5) The matrix $\mathbf{A}$ for GCN is constructed by pairwise cosine similarity between posts and sessions.

\begin{table*}[!t]
    \caption{The main performance comparison in four metrics for cyberbullying detection on two datasets. Note that the best model and the second model are highlighted by \textbf{bold} and \underline{underline}, respectively.}
    \centering   
    \begin{tabular}{c|c|c|c|c|c|c|c|c|}
    \hline
    Datasets & Metrics & CONcISE & RNN & GRU & GRU+A & \begin{tabular}[c]{@{}c@{}}LR\end{tabular} & \begin{tabular}[c]{@{}c@{}}RF\end{tabular} & HENIN \\ \hline \hline
    \multirow{4}{*}{Instagram} & Acc & 0.627 &0.782 & 0.815 & \underline{0.884} & 0.840 & 0.805 & \textbf{0.902} \\ \cline{2-9} 
    & Pre & 0.388 & 0.817 & 0.846 & 0.835 & 0.792 & \textbf{0.901} & \underline{0.889} \\ \cline{2-9} 
    & Rec & 0.381 & 0.376 & 0.496 & \underline{0.781} & 0.652 & 0.405 & \textbf{0.829} \\ \cline{2-9} 
    & F1 & 0.384 & 0.507 & 0.569 & \underline{0.805} & 0.715 & 0.559 & \textbf{0.838} \\ \hline
    \multirow{4}{*}{Vine} & Acc & 0.603 & 0.706 & 0.747 & \underline{0.797} & 0.788 & 0.786 & \textbf{0.804} \\ \cline{2-9} 
    & Pre & 0.363 & \textbf{0.830} & 0.773 & 0.757 & 0.748 & 0.751 & \underline{0.821} \\ \cline{2-9} 
    & Rec & 0.376 & 0.190 & 0.309 & \underline{0.559} & 0.512 & 0.498 & \textbf{0.643} \\ \cline{2-9} 
    & F1 & 0.369 & 0.245 & 0.418 & \underline{0.636} & 0.608 & 0.597 & \textbf{0.676} \\ \hline
    \end{tabular}
    \label{tab:model-performance}
\end{table*}

\subsection{Cyberbullying Detection Performance}
To answer \textbf{EQ1}, we first compare our HENIN with baseline methods. To evaluate the performance of cyberbullying detection methods, we use the following metrics, which are commonly used to evaluate classifiers: Accuracy (Acc), Precision (Pre), Recall (Rec), and F1-Score (F1). To have the experiments be more robust and reliable, we randomly choose $80\%$ of media sessions for training and the remaining $20\%$ for testing. We repeat the process $5$ times, and report the average values. The results are shown in Table~\ref{tab:model-performance}. We can find that the proposed HENIN consistently outperforms the competing methods across two datasets on Accuracy, Recall, and F1, i.e., except for the metric of Precision. 
Although RF and RNN lead to higher scores in Precision in Instagram and Vince datasets, respectively, their performance in other metrics is not stable. It is also worthwhile to notice that models considering attention mechanisms, i.e., HENIN and GRU+A, tend to produce better performance. This implies the importance of modeling contextual correlation and contribution at either word or sentence level on the detection of cyberbullying.


\subsection{Ablation Analysis for HENIN}
\label{sec:ablation}

\begin{figure}[!t]
    \centering
    \subfloat[Instagram]{{\includegraphics[width=0.5\linewidth]{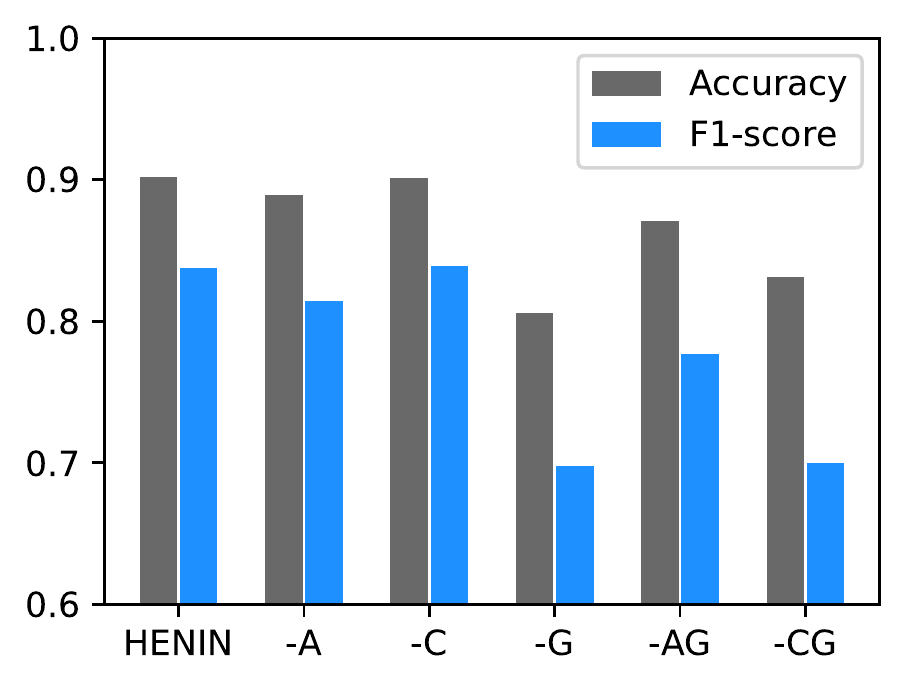} }}%
    \subfloat[Vine]{{\includegraphics[width=0.5\linewidth]{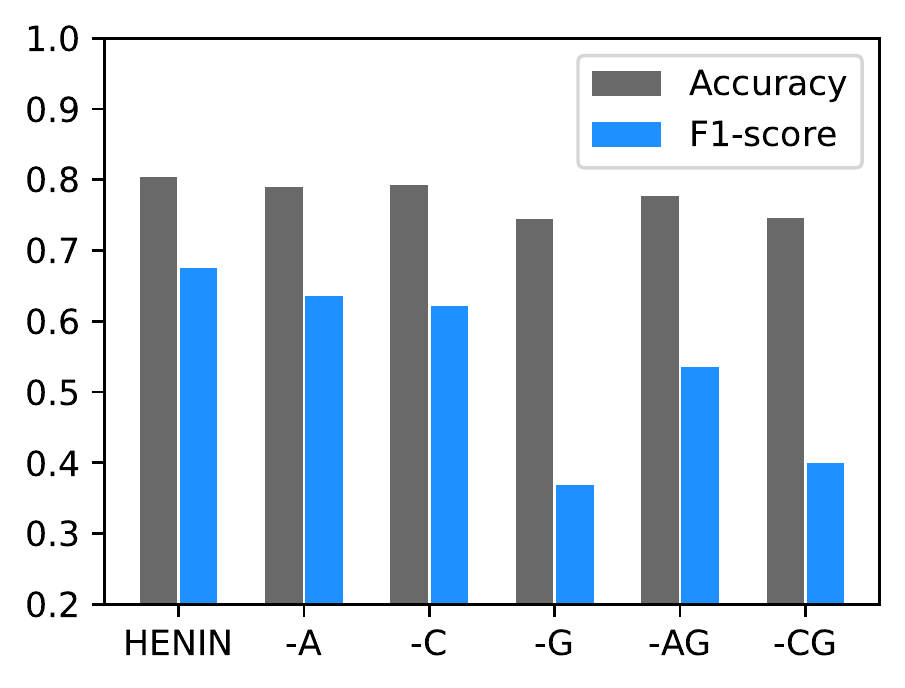} }}%
    \caption{Results of ablation analysis for HENIN.}%
    \label{fig:ablation-study}%
\end{figure}

To answer \textbf{EQ2}, we further investigate the effect of each component in the proposed HENIN model. We aim at evaluating the following reduced variants of HENIN.
(1) \textbf{-A}: HENIN without the Post-Comment co-attention component, (2) \textbf{-G}: HENIN without the GCN components, (3) \textbf{-C}: HENIN without the Comment Encoder, (4) \textbf{-AG}: HENIN without the Post-Comment co-attention and GCN components, and (5) \textbf{-CG}: HENIN without the Comment Encoder and GCN components.

The results are shown in Figure~\ref{fig:ablation-study}. The ablation analysis of HENIN brings two insights. First, all of the three components (i.e., comment encoder, session-session and post-post interactions, and posted text-comment co-attention) contribute apparently to the performance improvement. Second, When the model without considering the representations learned from session and post interactions, the performance reduces $14\%$ and $9.6\%$ in terms of F1-Score and Accuracy metrics on Instagram, and $30.7\%$ and $6\%$ on Vine. In other words, ``-G'' models hurt the performance most. The results suggest that modeling interactions between sessions and between posts through GCNs in HENIN is important.


\subsection{Early Detection of Cyberbullying}

\begin{figure}[!t]
    \centering
    \subfloat[Precision@10]{{\includegraphics[width=0.5\linewidth]{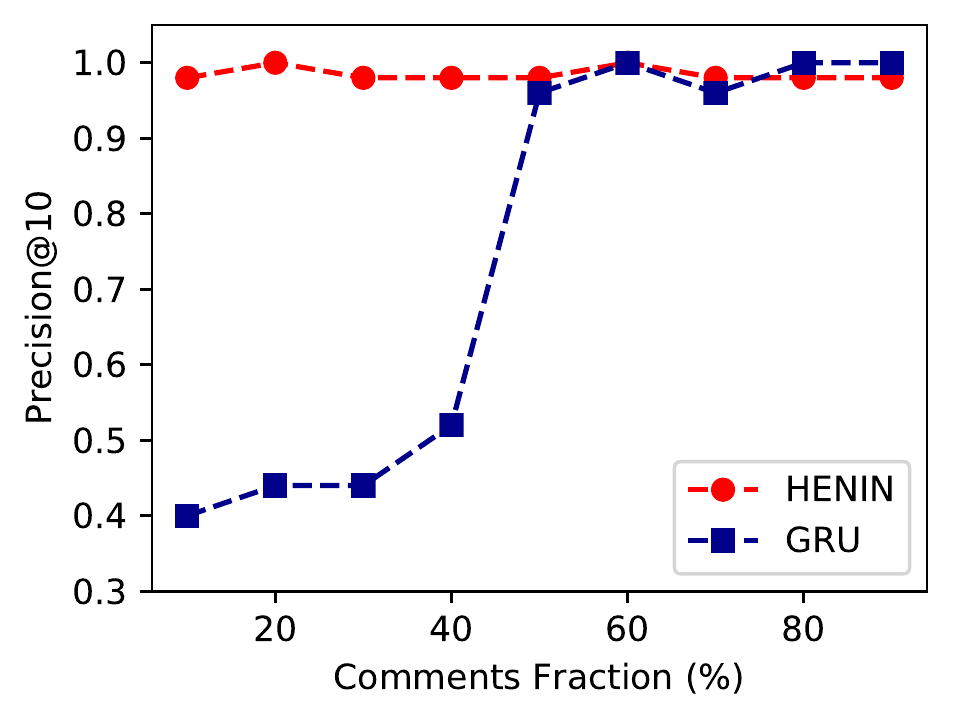} }}%
    \subfloat[Accuracy]{{\includegraphics[width=0.5\linewidth]{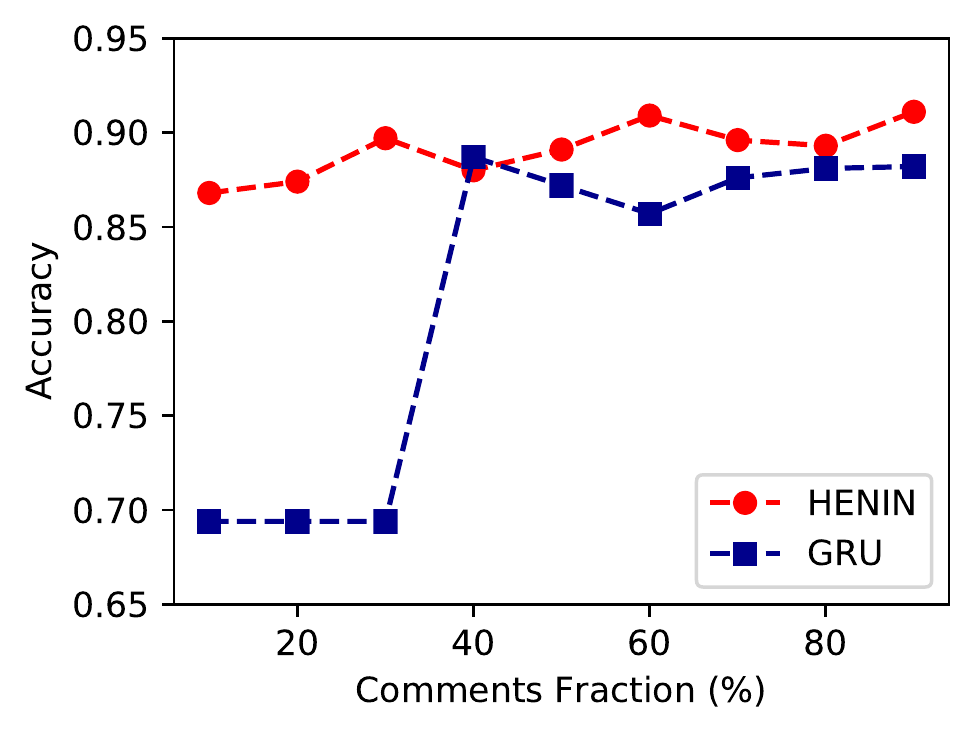} }}%
    \caption{Effect of comments' fraction on Instagram.}%
    \label{fig:commentFrac-insta}%
    \vspace{-4pt}
\end{figure}

\begin{figure}[!t]
    \centering
    \subfloat[Precision@10]{{\includegraphics[width=0.5\linewidth]{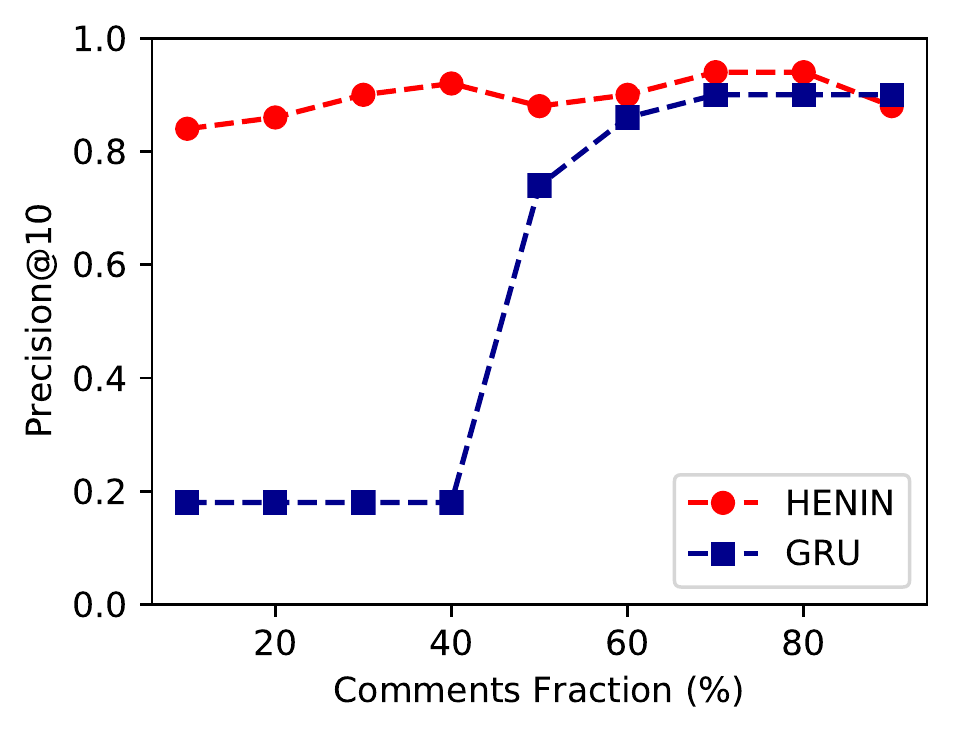} }}%
    \subfloat[Accuracy]{{\includegraphics[width=0.5\linewidth]{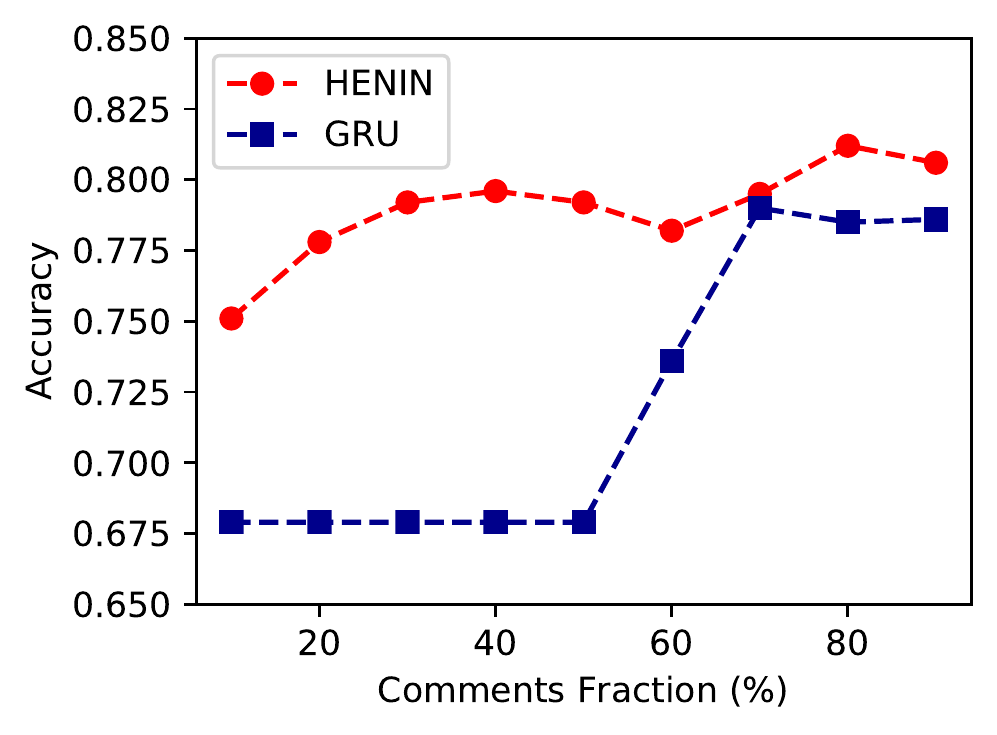} }}%
    \caption{Effect of comments' fraction on Vine.}%
    \label{fig:commentFrac-vine}%
    \vspace{-4pt}
\end{figure}

To answer \textbf{EQ3}, we examine whether HENIN can accurately detect cyberbullying sessions at early stages. In other words, we aim to understand how a model performs given only a partial proportion of observed comments. Here we choose GRU as the baseline for comparison. Specifically, for each media session, we sort all comments by response time, then choose various fractions of comments into the training and testing sets. We utilize \textit{Precision@k} and \textit{Accuracy} as the evaluation metrics, where $k=10$. The results are shown in Figure~\ref{fig:commentFrac-insta} and Figure~\ref{fig:commentFrac-vine}. From the figures, we can see that, our proposed HENIN can achieve much better performance when the observed comments are quite a few (i.e., the fraction of comments is low than $40\%$). In contrast, GRU model needs at least $50\%$ comments on both datasets to obtain the same good performance as HENIN. In short, we prove that HENIN is able to produce quite accurate early detection of cyberbullying sessions.

\subsection{Explainability and Case Study}
\paragraph{Explainability.} To answer \textbf{EQ4}, we evaluate the performance of the explainability of our \textit{HENIN} model from the perspective of comments. We choose \textit{GRU+A} as the baselines for comment explainability since it can learn attention weights for comments as a kind of explainability. Specifically, we want to see if the top-ranked explainable comments determined by our HENIN are more likely to be related to the major contexts in cyberbullying media sessions. We randomly choose $10$ media sessions, which contains at least $20$ but not more than $50$ comments, to evaluate the explainability ranking list of the comment \textit{RC}. Then we denote the ground-truth ranking list by rating the explainability score from $\{0, 1, 2, 3, 4\}$ for each comment, where $0$ means ``not explainable at all'', $1$ means ``not explainable'', $2$ means ``neutral'', $3$ means ``somewhat explainable'', and $4$ means ``highly explainable (highly malicious).'' We invite three domain experts to perform the ground-truth ratings for every comment. The average rating scores are used to generate the ranking list. Therefore, for each media session, we have two lists of top-$k$ comments, $L^{(1)}=\{L^{(1)}_1, L^{(1)}_2, ..., L^{(1)}_k\}$ by HENIN, and $L^{(2)}=\{L^{(2)}_1, L^{(2)}_2, ..., L^{(2)}_k\}$ by GRU+A. The top-$k$ comments are ranked and selected using the comment attention weights from high to low. To estimate the rank-aware explainability of comments, we utilize \textit{Normalized Discounted Cumulative Gain} (NDCG)~\citep{jarvelin2002cumulated} and \textit{Precision@k} as the evaluation metrics. We empirically set $k=10$. 

\begin{figure}[!t]
    \centering
    \subfloat[Precision@10]{{\includegraphics[width=0.5\linewidth]{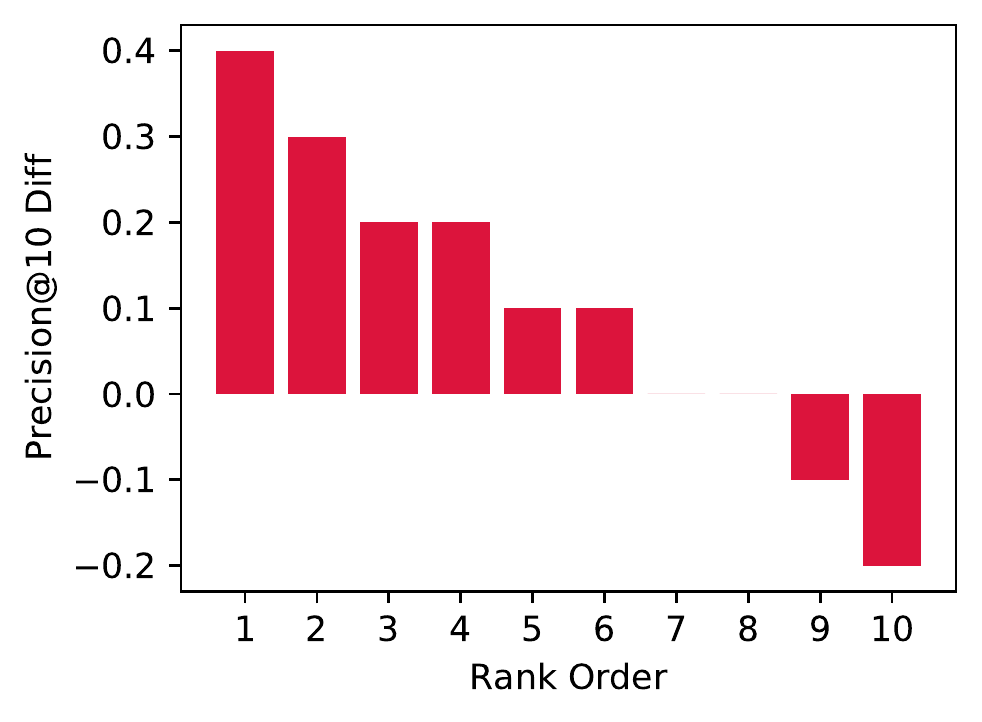} }}%
    \subfloat[NDCG@10]{{\includegraphics[width=0.5\linewidth]{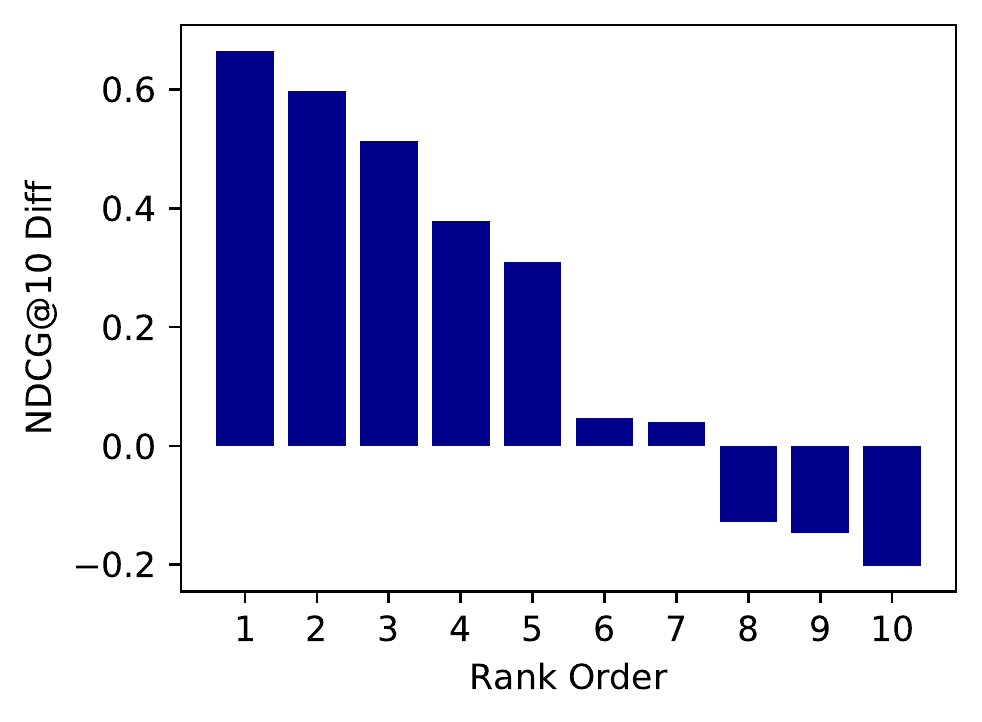} }}%
    \caption{The discrepancy histograms of mean Precision@10 and mean NDCG@10 (in the y-axis) for the results between HENIN and GRU+A in Vine dataset.}%
    \label{fig:discrepancy-histo}%
    \vspace{-12pt}
\end{figure}

The results are shown in Figure~\ref{fig:discrepancy-histo}, where media sessions are sorted by the discrepancy in the metrics between two methods, i.e., NDCG@$k$(HENIN)$-$NDCG@$k$(GRU+A), in a descending order. From the figures, we can have two observations. First, among $10$ Vine media sessions, HENIN obtains higher precision scores than GRU+A for $6$ cases. The overall mean precision scores over $10$ cases for HENIN and GRU+A are $0.51$ and $0.41$, respectively. Second, similar results can be found on NDCG scores. HENIN is superior to GRU+A on $7$ cases, and two cases have equal NDCG scores. The overall mean NDCG scores over $10$ cases for HENIN and GRU+A are $0.57$ and $0.36$, respectively. These results demonstrate that the attention weights of HENIN are able to highlight more evidential comments than GRU+A, and its explainability can be verified.

\begin{figure}[!t]
    \centering
    \includegraphics[width=1.0\linewidth]{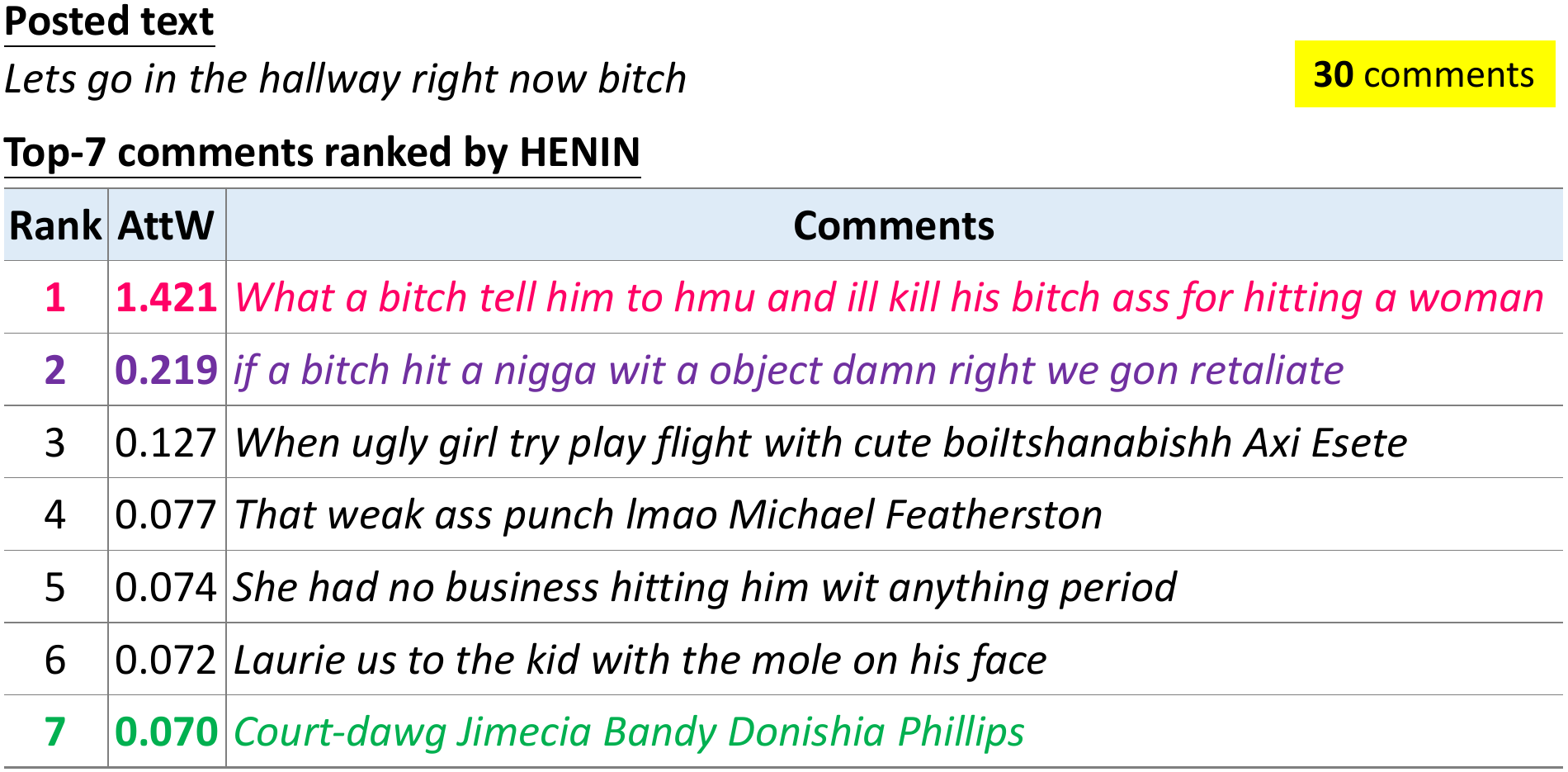}
    \caption{The top-$7$ comments highlighted by HENIN.}
    \label{fig:case-study}
    \vspace{-12pt}
\end{figure}

\paragraph{Case Study.} We further demonstrate the explainable comments that HENIN correctly ranks high but GRU+A misses. These cases are presented in Figure~\ref{fig:case-study}. We can find that: (1) our HENIN can rank more evidential comments higher than non-explainable comments. For example, the top-$1$ comment \textsl{``What a bitch tell him to hmu and ill kill his bitch ass for hitting a woman''} contains explicit vulgar and malicious texts that can explain why this media session detected as cyberbullying. (2) We can give higher attention weights to explainable comments than those neutral and unrelated comments. For example, the unrelated comment \textsl{``Court-dawg  Jimecia Bandy  Donishia Phillips''} has an attention weight $0.070$, which is lower than an explainable comment \textsl{``if a bitch hit a nigga wit a object damn right we gon retaliate''} with attention weight $0.219$. Therefore, the latter comment is selected to be a more important evidence for cyberbullying prediction. In short, HENIN is able to not only accurately detect cyberbullying sessions, but also highlight evidential comments as explanations.

\subsection{HENIN Hyperparameter Analysis}
Since we have shown that the graph-based interactions between sessions and between posts have a great impact on the detection (Section~\ref{sec:ablation}), we further aim to investigate how different hyperparameters of GCNs affect the performance. Here we study two hyperparameters. One is the number of GCN layers. The other is the choice of similarity measures in constructing the matrix $\mathbf{A}$ for GCN. The results on stacking the different number of GCN layers are shown in Table~\ref{tab:gcn-layers}. We can see that stacking more GCN layers leads to performance improvement by around $1.1\%$ in terms of F1 on Instagram and $2.2\%$ on Vine.

\begin{table}[!t]
    \caption{Effect of the number of GCN layers.}
    \centering
    \begin{tabular}{lcccc}
    \hline 
    Dataset & \multicolumn{2}{c}{Instagram} & \multicolumn{2}{c}{Vine} \\ \hline
    & Acc & F1 & Acc & F1 \\ \hline 
    \#layers=1 & 0.896 & 0.827 & 0.803 & 0.672 \\
    \#layers=2 & 0.896 & 0.829 & 0.797 & 0.654 \\
    \#layers=3 & \textbf{0.902} & \textbf{0.838} & \textbf{0.804} & \textbf{0.676} \\ \hline
    \end{tabular}
    \label{tab:gcn-layers}
\end{table}

\begin{table}[!t]
    \caption{Effect of similarity measures in constructing matrix $\mathbf{A}$ depicting the graph for GCN.}
    \centering
    \begin{tabular}{ccccc}
    \hline 
    Dataset & \multicolumn{2}{c}{Instagram} & \multicolumn{2}{c}{Vine} \\ \hline
    $\mathbf{A}_{ij}$& Acc & F1 & Acc & F1 \\ \hline 
    $\text{cos}(\mathbf{x}_i, \mathbf{x}_j)$ & 0.894 & 0.823 & 0.806 & 0.668 \\
    $\text{jac}(\mathbf{x}_i, \mathbf{x}_j)$ & 0.893 & 0.824 & \textbf{0.811} & \textbf{0.673} \\
    $\text{euc}(\mathbf{x}_i, \mathbf{x}_j)$ & \textbf{0.922} & \textbf{0.872} & 0.794 & 0.661 \\ \hline
    \end{tabular}
    \label{tab:gcn-adjacency}
\end{table}

The weight matrix $\mathbf{A}$ for GCN is obtained by calculating the similarity for all pairs of nodes in the graph. We compare three commonly similarity measures, Cosine similarity: $\text{cos}(\mathbf{x}_i, \mathbf{x}_j)=\frac{\mathbf{x}_i\cdot \mathbf{x}_j}{\left \| \mathbf{x}_i \right \|\left \| \mathbf{x}_j \right \|}$, Jaccard similarity: $\text{jac}(\mathbf{x}_i, \mathbf{x}_j)=\frac{\mathbf{x}_i\mathbf{x}_j}{\sum \mathbf{x}_i\sum \mathbf{x}_j-\sum \mathbf{x}_i \mathbf{x}_j}$, and Euclidean similarity: $\text{euc}=1-\bar{\text{euc}}(\mathbf{x}_i, \mathbf{x}_j)=1-\bar{N}(\sqrt{\sum (\mathbf{x}_i-\mathbf{x}_j)^2}$) ($\bar{\text{euc}}$ and $\bar{N}$ denote normalization to [0,1]). The results are shown in Table~\ref{tab:gcn-adjacency}. We can see that on the Instagram dataset, using Euclidean similarity can improve the performance by $4.9\%$ and $2.8\%$ in terms of F1 and Accuracy, respectively. On the Vine dataset, using Jaccard similarity outperform than the other two measures by improving $1.2\%$ and $1.7\%$ in terms of F1 and Accuracy, respectively. The results suggest that in different datasets, we need to choose the proper similarity measure to construct the weight matrix as the performance can be affected.

\section{Conclusion}
Cyberbullying detection on social media attracts growing attention in recent years. It is also crucial to understand why a media session is detected as cyberbullying. Thus we study the novel problem of explainable cyberbullying detection that aims at improving detection performance and highlighting explainable comments. We propose a novel deep learning-based model, HEterogeneous Neural Interaction Networks (HENIN), to learn various feature representations from comment encodings, post-comment co-attention, and graph-based interactions between sessions and posts. Experimental results exhibit both promising performance and evidential explanation of HENIN. We also find that the learning of graph-based session-session and post-post interactions contributes most to the performance. Such results can encourage future studies to develop advanced graph neural networks in better representing the interactions between heterogeneous information. In addition, it is worthwhile to further model information propagation and temporal correlation of comments in the future.

\section*{Acknowledgments}
This work is supported by Ministry of Science and Technology (MOST) of Taiwan under grants 109-2636-E-006-017 (MOST Young Scholar Fellowship) and 109-2221-E-006-173, and also by Academia Sinica under grant AS-TP-107-M05.

\bibliographystyle{acl_natbib}
\bibliography{acl2020}

\end{document}